\pdfoutput=1

\documentclass[a4paper, 10pt, conference]{ieeeconf}      % Use this line for a4 paper

\usepackage{blindtext, graphicx}
\usepackage{gensymb}
\usepackage{xcolor}

\usepackage{tikz,lipsum}
\usepackage[labelformat=simple]{subcaption}

\DeclareCaptionLabelSeparator{periodspace}{.\quad}
\usepackage{listings}

\usepackage{algorithm}
\usepackage{algorithmic}

\usepackage{ amssymb }
\usepackage{amsmath}
\usepackage{commath}

\usepackage{framed} 

\usepackage{float}

\usepackage{array}
\newcolumntype{P}[1]{>{\centering\arraybackslash}p{#1}}
\newcolumntype{M}[1]{>{\centering\arraybackslash}m{#1}}

\newcommand*{\prob}{\mathsf{P}}

\IEEEoverridecommandlockouts                              % This command is only needed if 
\title{\LARGE \bf
Transfer Learning for Unseen Robot Detection and Joint Estimation on a Multi-Objective Convolutional Neural Network
}

\author{Justinas Mi\v{s}eikis$^{1}$, Inka Brijacak$^{2}$, Saeed Yahyanejad$^{3}$, Kyrre Glette$^{4}$, Ole Jakob Elle$^{5}$, Jim Torresen$^{6}$% <-this % stops a space
\thanks{$^{1}$ $^{4}$ $^{5}$ $^{6}$Justinas Mi\v{s}eikis, Kyrre Glette, Ole Jakob Elle and Jim Torresen are with the Department of Informatics, University of Oslo, Oslo, Norway}
\thanks{$^{2}$ $^{3}$ Inka Brijacak and Saeed Yahyanejad are with the Joanneum Research - Robotics, Klagenfurt am W\"orthersee, Austria} 
\thanks{$^{5}$Ole Jakob Elle has his main affiliation with The Intervention Centre, Oslo University Hospital, Oslo, Norway {\tt\small oelle@ous-hf.no}}%
\thanks{$^{1}$ $^{4}$ $^{6}$ {\tt\small \{justinm,kyrrehg,jimtoer\}@ifi.uio.no}}%
\thanks{$^{2}$ {\tt\small Inka.Brijacak@joanneum.at}}%
\thanks{$^{3}$ {\tt\small Saeed.Yahyanejad@joanneum.at}}%
}

\begin{document}

\maketitle
\thispagestyle{empty}
\pagestyle{empty}

%%%%%%%%%%%%%%%%%%%%%%%%%%%%%%%%%%%%%%%%%%%%%%%%%%%%%%%%%%%%%%%%%%%%%%%%%%%%%%%%
\begin{abstract}

A significant problem of using deep learning techniques is the limited amount of data available for training. There are some datasets available for the popular problems like item recognition and classification or self-driving cars, however, it is very limited for the industrial robotics field. In previous work, we have trained a multi-objective Convolutional Neural Network (CNN) to identify the robot body in the image and estimate 3D positions of the joints by using just a 2D image, but it was limited to a range of robots produced by Universal Robots (UR). In this work, we extend our method to work with a new robot arm - Kuka LBR iiwa, which has a significantly different appearance and an additional joint. However, instead of collecting large datasets once again, we collect a number of smaller datasets containing a few hundred frames each and use transfer learning techniques on the CNN trained on UR robots to adapt it to a new robot having different shapes and visual features. We have proven that transfer learning is not only applicable in this field, but it requires smaller well-prepared training datasets, trains significantly faster and reaches similar accuracy compared to the original method, even improving it on some aspects.

\end{abstract}

%%% TO DO %%%

% Forward propagation time - 
% Figure with successful detections of robot joints on Kuka - have it vertical 2x3?
% 

\section{INTRODUCTION}

Industrial robotics has been associated with structured and well-defined environments for many years and robot arms have achieved great performance in areas like manufacturing. It comprises of hard-coded repetitive motions, where a machine can do a better job compared to a person in terms of no fatigue, precision and non-stop operation. However, with developing hardware, computing power and advancing algorithms, the same systems are becoming more adaptive. Nowadays, instead of fencing off the robots, environment understanding and adaptive behaviour is a part of the Industry 4.0 concept, where robots and people can share the same workspace and collaborate~\cite{lee2015cyber}.

There are numerous approaches to sense the environment: laser scanners, stereo vision, RGB-D cameras, camera arrays, ultrasound sensors, motion capture systems. Each one has its own pros and cons, often either needing additional markers or calibrated devices or having a high price-tag. Very often there is still a significant amount of work needed to set up a new robustly working system.

Inspiration of the environment understanding comes from biology - how animals and especially humans are able to understand the environment. We are capable of learning what objects are, how they move, their functionality and the way we should interact with them by looking at example situations. Furthermore, after we know how it works in some situation, it is very likely that next time we see similar conditions, we will be able to find parallels between the two and figure out how we should act by simply using our previously gained knowledge. That is the motivation of the transfer learning method, which uses a previous well-trained neural network and adjusts it to new conditions using limited amount of training data and significantly shorter training time compared to the full training of the neural network.

\begin{figure*}[ht]
\vspace{0.2cm}
\centering
    \includegraphics[width=0.99\linewidth]{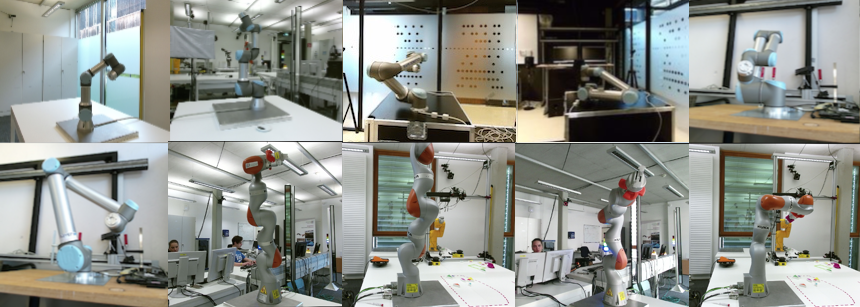}
    \caption{Samples from the collected robot datasets. It consists of a line of Universal Robots (silver-blue) as well as Kuka LBR iiwa (silver-orange). The data was collected with a variety of backgrounds and light conditions to provide more robustness.}
    \label{fig:dataset_example_images}
\vspace{-0.5cm}
\end{figure*}

Transfer learning has been used in a variety of fields. In many cases, the whole or part of the CNN trained on ImageNet is taken as a base network and then adjusted to a specific application~\cite{krizhevsky2012imagenet}. This has been proven to work for mid-level image representations in object classification, using the pre-trained network on natural images to adapt for medical image recognition~ and even emotion recognition~\cite{oquab2014learning}~\cite{greenspan2016guest}~\cite{ng2015deep}. Another interesting application of transfer learning is to use a fully trained network on night-time satellite imagery of poverty areas and adapt it to recognise poverty areas from daytime satellite imagery~\cite{xie2015transfer}. Furthermore, detailed analyses of the transfer learning approaches were made with surveys of the techniques used and various CNN structures~\cite{shin2016deep}~\cite{weiss2016survey}.

The proof that generalised visual features can be transferred to new systems has motivated to use it to extend our previous work of recognising the robot and estimating its 3D position of the joints by using a simple 2D color camera image~\cite{miseikis2018multi}. Instead of using ImageNet or any other well known pre-trained network, we take our previously fully trained multi-objective CNN on Universal Robots and use it to adapt to a new Kuka LBR iiwa robot arm. Additionally, the new dataset adds new unseen backgrounds making the network even more robust.

The main goal of identifying the robot in a 2D camera image is to remove the need for fully calibrated camera-robot systems allowing for more dynamic environments, while still ensuring safe operations. It is crucial for shared workspaces between humans and robots. There are many good methods of real-time dynamic obstacle and people avoidance, but most of them require a fully-calibrated robot-camera system~\cite{mainprice2013human}~\cite{miseikis2016multi}. Despite some efficient Hand-Eye calibration methods, it is still a cumbersome process when the operation of the robot has to be halted until the calibration is completed~\cite{miseikis2016automatic}. Furthermore, it can simplify the task of having mobile robots moving around the floor without any special markings. By identifying other fixed robots it can both avoid possible collisions and localise itself to known fixed-base robots. By identifying other robots and knowing their exact position, the setup could be expanded to prediction of the behaviour of other machinery in the surrounding environment without having the direct communication channel between them. This would be a very useful approach in swarm robotic applications.

This paper is organized as follows. We present the system setup and dataset collection in Section~\ref{sec:system_setup}. Then, we explain the proposed method and CNN structure and configuration in Section~\ref{sec:method} and the transfer learning procedure in Section~\ref{sec:training}. We provide experiments and results in Section~\ref{sec:results}, followed by relevant conclusions and future work in Section~\ref{sec:conclusion}.

% Section System Setup
\section{SYSTEM SETUP AND DATASET COLLECTION}
\label{sec:system_setup}

Training a deep learning network typically requires a large amount of diverse training data. The main problem lies in the necessity to have precise ground-truth information, which is given as a correct answer.

Our setup consisted of a vision sensor, in this case, a Kinect V2 camera, placed in arbitrary positions overlooking the robot and perform Hand-Eye calibration at each of the positions~\cite{Fankhauser2015KinectV2ForMobileRobotNavigation}. The calibration is done by placing a known marker on the end-effector of the robot and performing a number of movements until the calibration accuracy reaches the necessary precision. The result is a coordinate frame transformation between the camera and the robot base~\cite{heikkila2000flexible}. 
%This method works well, but the limiting factor is that during the recording, the camera position has to be fixed. If the camera or the robot is moved relative to each other, the calibration process has to be repeated.

%Another workaround is to make use of a motion capture (mocap) system for tracking a relative position between the camera and the robot. Mocap system works by placing an infrared (IR) reflective markers on the objects of interest and having multiple IR cameras placed around the room to precisely track the objects. An Optitrack mocap system was used in the project and given a good visibility of each object it has the capability of tracking each object with sub-millimeter accuracy~\cite{point2011optitrack}.

%{\color{red} ADD MOCAP SETUP IMAGE }

Given a precise coordinate frame transformation, the robot model is used together with the live information from its joint encoder readings to create a simplified mesh model defining the robot shape. Then it is transformed to the coordinate frame of the camera and depth image estimated from the viewpoints of the camera. The result is a precise mask of the robot body in the camera image, which can be overlayed with a color image and used as a ground truth data for teaching the CNN. The main benefit is that this process is fully automated by using ROS with MoveIt! package~\cite{sucan2013moveit}. The robot model is taken from the Unified Robot Description Format (URDF) files provided by the robot manufacturers~\cite{meeussen2012urdf}.

In our experiments, we use an already trained multi-objective CNN from the previous project~\cite{miseikis2018multi}, which was trained from scratch on three robot models from Universal Robots: UR3, UR5 and UR10. In order to test the capabilities of transfer learning, new datasets using Kuka LBR iiwa were used. For comparison reasons, relatively large datasets, summarised in Table~\ref{table:dataset_summary_new}, were collected for all the robots. These datasets consist of multiple recordings, each one with the camera placed at different angles and distances relative to the robot as well as having various backgrounds.

\begin{table}[h]
\caption{Dataset summary describing a number of samples collected for each type of the robot. }
\label{table:dataset_summary_new}
\centering
\begin{tabular}{ |M{2cm}||M{2.4cm}|M{2.4cm}|}
 \hline
 Robot Type & Number of Datasets & Total Number of Samples \\
 \hline
 Universal Robots & $9$ & $4350$ \\
 Kuka LBR iiwa & $14$ & $1837$ \\
% Franka Emika Panda Robot & xxx\\
 \hline
\end{tabular}
%\vspace{-0.2cm}
\end{table}

Robot movements included a large variety of joint configurations resulting in many viewpoints of the robot. Furthermore, lighting conditions were varied for each of the recordings to allow for more robustness regarding the brightness and reflections.

The new datasets with the Kuka robot also included more dynamic background with people moving around and even another Kuka robot placed further away and not being used in experiments. Furthermore, in some cases, the robot went out of bounds of the color image. In total, 9 datasets of Universal Robots and 14 datasets of Kuka robot were used. Each recording had different camera placement, changing distance between the robot and the camera, varying lighting conditions and new background. During each of the recordings, the robot was moving to give a large variety of joint configurations in the dataset.

%%% ADD THE MASK, JOINT COORDS AND BASE POS AS GROUND TRUTH %%%
\begin{figure}[ht]
\vspace{0.2cm}
\centering
\begin{subfigure}[t]{0.23\textwidth}
    \includegraphics[width=\textwidth]{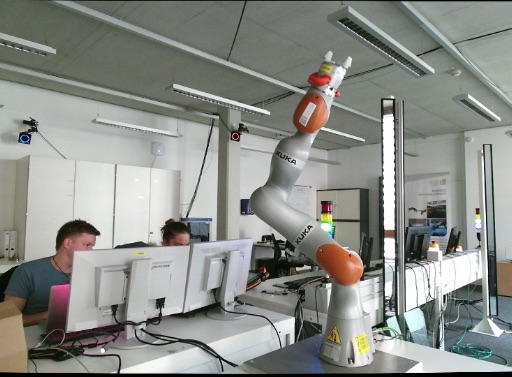}
    \caption{Color image from the dataset used as an input.}
    \label{fig:input_color_image}
\end{subfigure}
~
\begin{subfigure}[t]{0.23\textwidth}
    \includegraphics[width=\textwidth]{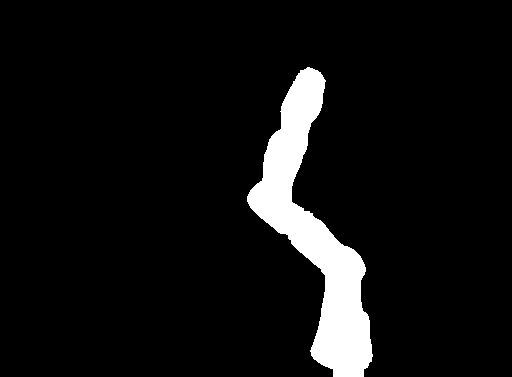}
    \caption{Ground truth model of the robot mask.}
    \label{fig:input_gt_image}
\end{subfigure}
~
\begin{subfigure}[t]{0.23\textwidth}
    \includegraphics[width=\textwidth]{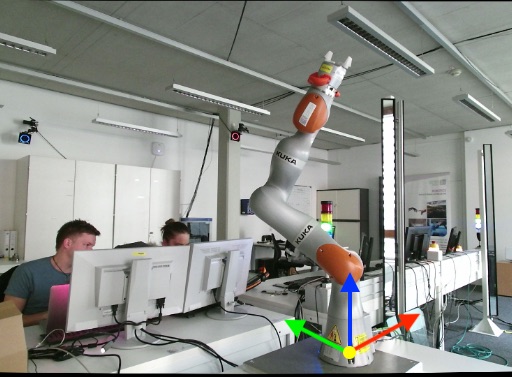}
    \caption{Ground truth data of the robot base 3D position.}
    \label{fig:input_gt_robot_base}
\end{subfigure}
~
\begin{subfigure}[t]{0.23\textwidth}
    \includegraphics[width=\textwidth]{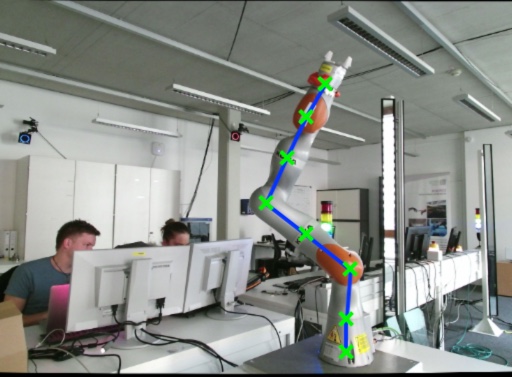}
    \caption{Ground truth data of the 3D position of robot joints.}
    \label{fig:input_gt_joints}
\end{subfigure}
\caption{Dataset, mask and ground truth value examples of the Kuka LBR iiwa robot.}
\label{fig:input_data}
\vspace{-0.5cm}
\end{figure}

At the completion of each movement, a trigger signal was sent in order to save the color image, depth model, cartesian and joint coordinates of each of the robot joint and ground-truth mask model of the robot. All this information was later used to train the neural network. However, depth information was used only for training, while the recognition part of the system relies only on the color camera image as an input.

In order to normalise the input data, internal camera calibration was used to ensure a perfect overlap between color and depth images. All the input images are also rectified and have the resolution of $512\times424$ pixels. Testing and validation sets were divided by the ratios of $80\%$ and $20\%$ respectively based on random sampling.

{\color{blue}

\begin{figure*}[ht]
\vspace{0.2cm}
    \centering
    \includegraphics[width=0.99\textwidth]{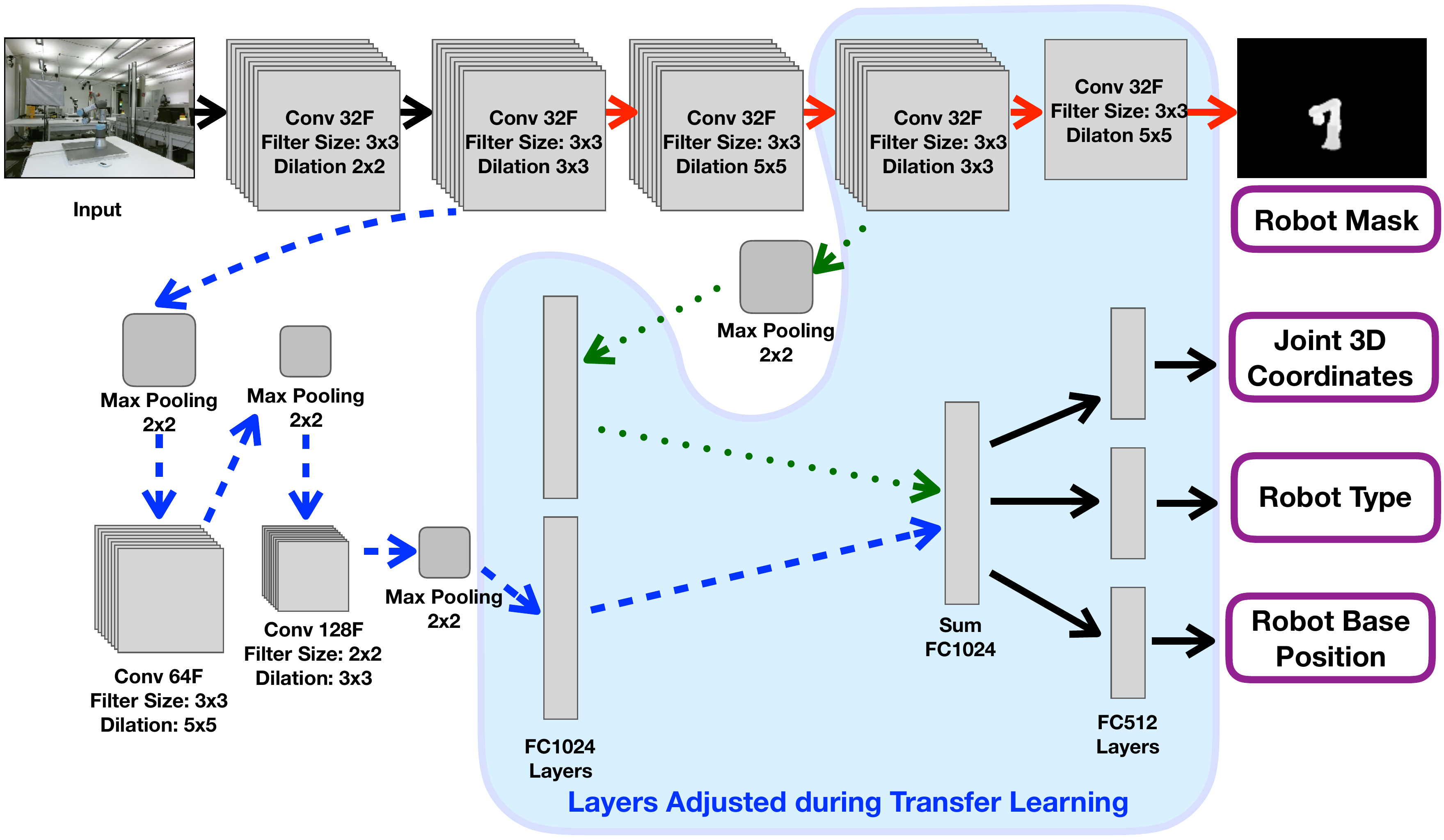}
    \caption{
    Structure of the multi-objective CNN. Input is a 2D color image resulting in four outputs: robot mask, 3D coordinates of robot joints, 3D coordinates of the robot base and the robot type. The network uses part of common convolutional layers and then branches off for objective-specific training. Fully-connected (FC) layers, marked by blue area, are the ones being adjusted during the transfer learning process to adapt to the new robot model. Convolutional layers learn generalised visual features of the image, so their parameters stay \textit{frozen} during the transfer learning. This allows for quicker adaptation with a limited number of input images compared to the full training process. The whole CNN is trained for all four outputs simultaneously using a common loss function. Differently colored arrows represent connections of different branches in multi-objective CNN, each one focused for a certain type of output.}
    \label{fig:multi_obj_cnn_structure}
\vspace{-0.5cm}
\end{figure*}

%Recorded images have $512\times424$ pixel resolution and they are all rectified to compensate for lens distortion. Internal camera calibration was used to ensure that both color and depth information have a perfect overlap, avoiding any offsets. Random sampling was used to divide the final dataset into the training set and the test set by ratios of $80\%$ and $20\%$ of all the images respectively.

}

% NEW Methods
\section{CNN STRUCTURE AND CONFIGURATION}
\label{sec:method}

The base of a multi-objective CNN is taken from previous work, where it was trained on a line of robots made by Universal Robot~\cite{miseikis2018multi}. The network simultaneously optimises for multiple heterogeneous outputs by using just a single image as an input.

The network in this paper is trained on four objectives:
\begin{itemize}
  \item Robot mask in the image
  \item Robot type
  \item 3D Robot base position in relation to the camera
  \item 3D Position of the robot joints
\end{itemize}

The structure of the CNN is shown in Figure~\ref{fig:multi_obj_cnn_structure}. The network shares a number of common convolutional layers and then branches for more objective-specific optimisation. Having a single training process, it means that the features in common layers are reused.

{\color{blue} 

}

\subsection{Loss Functions}

% Given the simultaneous training of the whole network for all four objectives, a common loss function has to be used. It actually consists of four separate loss functions, one for each goal, connected together. First, each loss function will be described separately followed by the explanation, how they are combined into a single one used in training.

Loss functions are used to evaluate the training progress and the achieved accuracy compared to the ground truth data. Our system optimises for four objectives simultaneously, resulting in four loss functions, which are later combined into one for the training process. First, each of the loss functions will be described separately followed by the explanation of how they are all connected into one.

The robot body takes up a relatively small area in the whole image. The area taken up by the robot body in UR datasets is between $6-17\%$ and for Kuka datasets, it is between $8-18\%$ of the whole image. Given a standard pixel classification loss function, there would be cases when an accuracy of over $82\%$ can be reached by classifying the whole image as a background. That is conceptually wrong, so the loss function was adjusted by using the foreground weight $w_{fg}$, which is calculated in Equation~\ref{eq:fg_weight}. It is based on the inverse probability of the foreground and background classes, where $Y \in \{fg, bg\}$.

\begin{equation}
    w_{fg} = \frac{1}{\prob(Y=fg)}
\label{eq:fg_weight}
\end{equation}

The background weight $w_{bg}$ is calculated in Equation~\ref{eq:bg_weight}.

\begin{equation}
    w_{bg} = \frac{1}{\prob(Y=bg)}
\label{eq:bg_weight}
\end{equation}

The loss function for the robot mask is defined by two steps. First, a per-pixel loss $l^n$ is calculated in Equation~\ref{eq:loss_classification_pixel}, where $i_{est}$ is $\prob(Y = fg)$, $(1-i_{est})$ is $\prob(Y = bg)$ and $i_{gt}$ is the ground truth value from the mask image.

\begin{equation} \label{eq:loss_classification_pixel}
    \begin{split}
        l^n (I_{est}^n, I_{gt}^n) = 
        & -w_{fg} i_{est} \log{(i_{gt})} \\
        & - w_{bg}(1-i_{est})\log{(1-i_{gt})}
    \end{split}
\end{equation}

This is followed by a normalised loss calculation for the whole image $\mathcal{L}_{mask}$ in Equation~\ref{eq:loss_classification_full}. A normalisation factor $\mathcal{N}$, which is the number of pixels in the image, allows us to keep the same learning parameters independent of the input image size. 
    
\begin{equation}  \label{eq:loss_classification_full}
    \mathcal{L}_{mask} (I_{est}, I_{gt}) = \frac{1}{\mathcal{N}} \sum\limits_{n} l^n (i_{est},  i_{gt})
\end{equation}

3D coordinates of the robot base and robot joints are defined as regression tasks. The loss function is based on the Euclidean distance between the estimated values and the ground truth values. For the robot joints estimation, the loss function $\mathcal{L}_{Jcoords}$ is described in Equation~\ref{eq:loss_joints_coords}, where $N_j$ is the number of joints, $J_{i}$ is the ground truth position of each joint and $E_{i}$ is the estimated values by the neural network.

\begin{equation} \label{eq:loss_joints_coords}
    \mathcal{L}_{Jcoords} = \frac{1}{N_j} \sum\limits_{i=1}^{N_j} \norm{J_{i}-E_{i}}_2
\end{equation}

The loss function for the coordinates of the robot base $\mathcal{L}_{Bcoords}$ is calculated in Equation~\ref{eq:loss_base_coords}. $B_{xyz}$ is the ground truth position of the robot base in 3D, and $E_{xyz}$ is the estimated 3D position of the robot base. These positions are relative to the coordinate frame of the camera.

\begin{equation} \label{eq:loss_base_coords}
    \mathcal{L}_{Bcoords} = \norm{B_{xyz}-E_{xyz}}_2
\end{equation}

% Robot type loss function
Classification of the robot type $\mathcal{L}_{type}$ is defined as a categorical cross-entropy problem with multiple classes. $\mathcal{L}_{type}$ is calculated in Equation~\ref{eq:loss_robot_type}, where $p$ is the ground truth labels, $q$ are the predicted labels and $c\in R$, where $R$ contains all the available types of robots in the dataset.

\begin{equation} \label{eq:loss_robot_type}
    \mathcal{L}_{type} = -\sum\limits_{c} p(c) \log{q(c)}
\end{equation}

% Combined loss function

For the training of the multi-objective CNN and optimisation for all four objectives, a single loss function is needed. This was achieved by combining the previously defined loss functions into $\mathcal{L}_{final}$ by having a weight element for each of the losses, as shown in Equation~\ref{eq:final_loss}. The larger the weight $W$, the higher the impact on the corresponding value.

\begin{equation} \label{eq:final_loss}
    \begin{split}
        \mathcal{L}_{final} = 
        & W_{mask}\mathcal{L}_{mask} + W_{Jcoords}\mathcal{L}_{Jcoords} \\
        &+ W_{Bcoords}\mathcal{L}_{Bcoords} + W_{type}\mathcal{L}_{type}
    \end{split}
\end{equation}

\section{TRANSFER LEARNING AND TRAINING}
\label{sec:training}

The benefit of transfer learning technique is that the parameters contained in so-called \textit{frozen} layers are copied from the previously trained network, while only part of layers is trained during the process. This speeds up the training process and requires smaller training datasets compared to the full CNN training. In this work, most of the convolutional layers had the parameters transferred and \textit{frozen} with all the fully connected layers and only the two last convolutional layers for robot mask estimation being trained to adapt for specific variation in visual features. They contain more robot-specific visual features, while the first layers learn more general visual features, which are more adaptable for any robot type. The exact setup is explained in Figure~\ref{fig:multi_obj_cnn_structure}. By a layer being \textit{frozen} it means that after the parameter transfer, they are fixed and not adjusted at all during the training.

Weights for the loss function are kept identical to the ones in previous work given good results and ability to compare the results of the works directly. Selected weight values were the following:
\begin{itemize}
    \item $W_{mask}$: $1.0$
    \item $W_{Jcoords}$: $1.5$
    \item $W_{Bcoords}$: $1.5$
    \item $W_{type}$: $0.3$
\end{itemize}

One important difference between the UR robots and the Kuka robot is the number of joints. Universal Robot line has 6 joints, while Kuka has 7 joints. This difference changes the number of outputs for the 3D position estimation of robot joints. However, because the fully connected layers, as well as output layers, are trained, it can be adjusted to accommodate estimation of an extra joint.

\begin{figure*}[ht]
\vspace{0.2cm}
    \hfill
    \centering
    \begin{subfigure}[t]{0.32\textwidth}
        \centering
        \includegraphics[width=\linewidth]{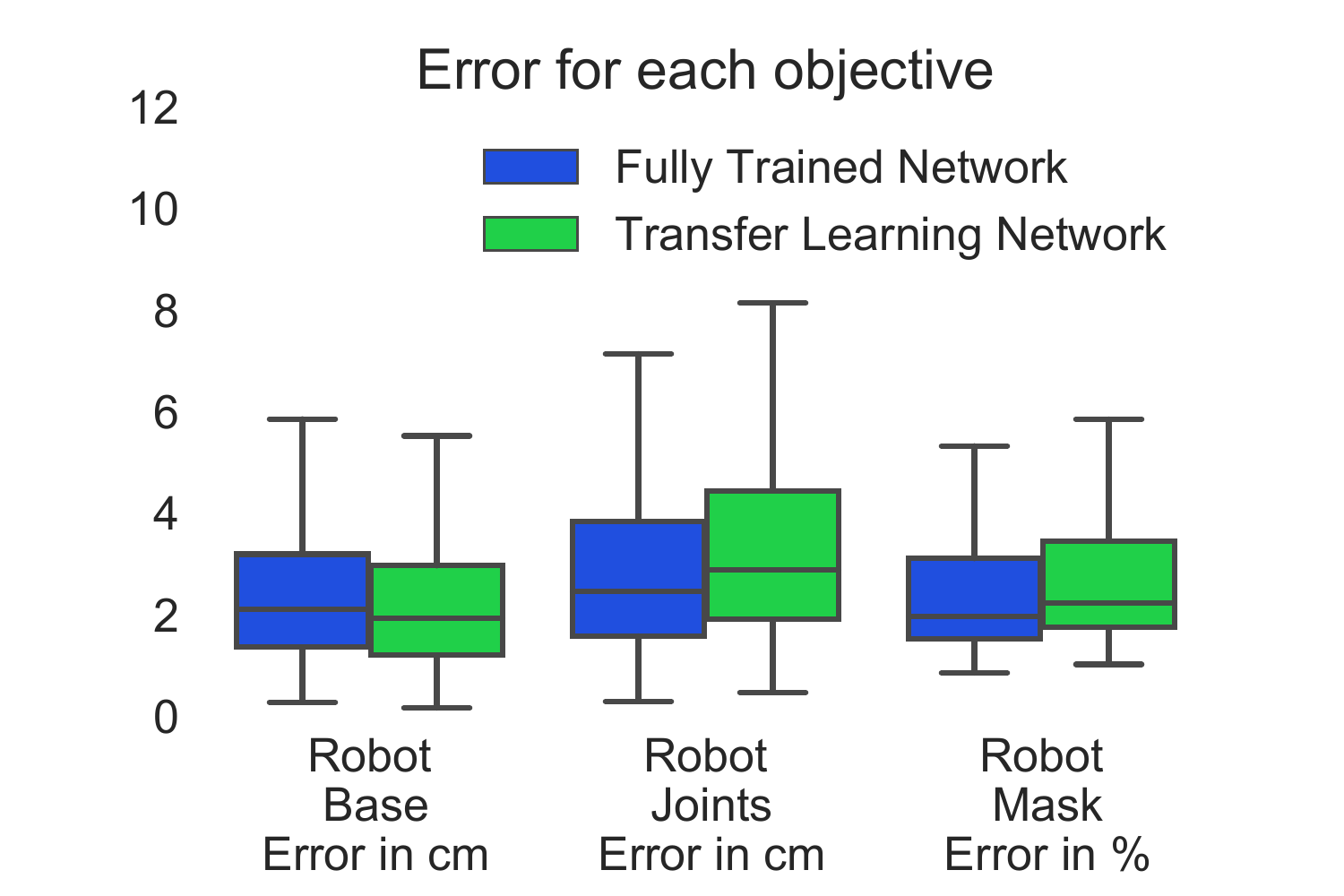}
        \caption{Comparison of a fully-trained network on UR robots~\cite{miseikis2018multi} with a transfer learning method on Kuka. Error slightly increased for joints position estimation and robot mask, however, transfer learning method was slightly more accurate in estimating the position of the robot base.}
        \label{fig:results_all} 
    \end{subfigure}
    \hfill
    \begin{subfigure}[t]{0.32\textwidth}
        \centering
        \includegraphics[width=\linewidth]{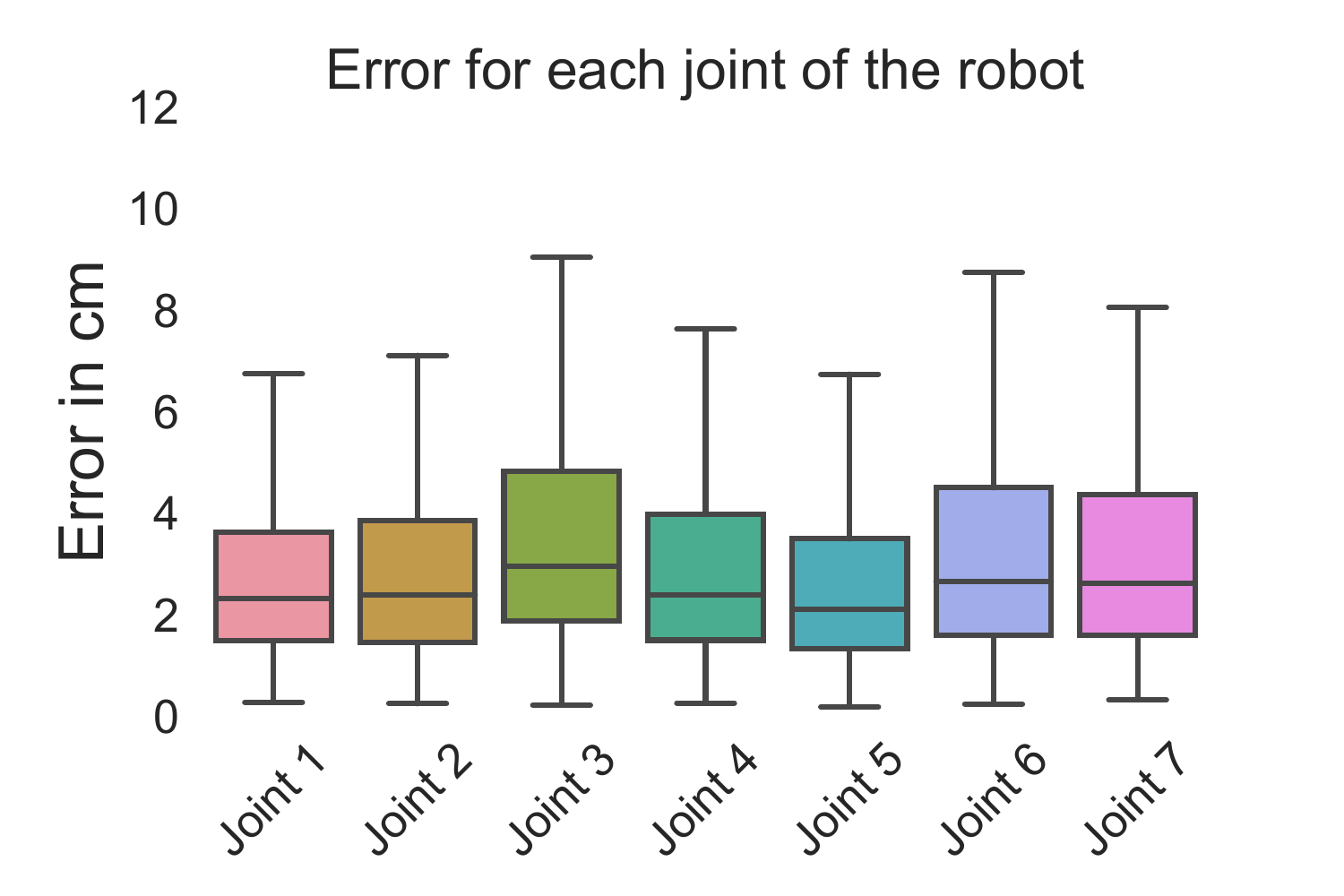}
        \caption{Error for the 3D coordinate estimation for positions of each robot joint of the transfer learning method. It can be seen that the error slightly increases for joints further away from the base and Joint 3 has increased error compared to the surrounding joints}
        \label{fig:results_joints}
    \end{subfigure}
    \hfill
    \begin{subfigure}[t]{0.32\textwidth}
        %\centering
        \includegraphics[width=\linewidth]{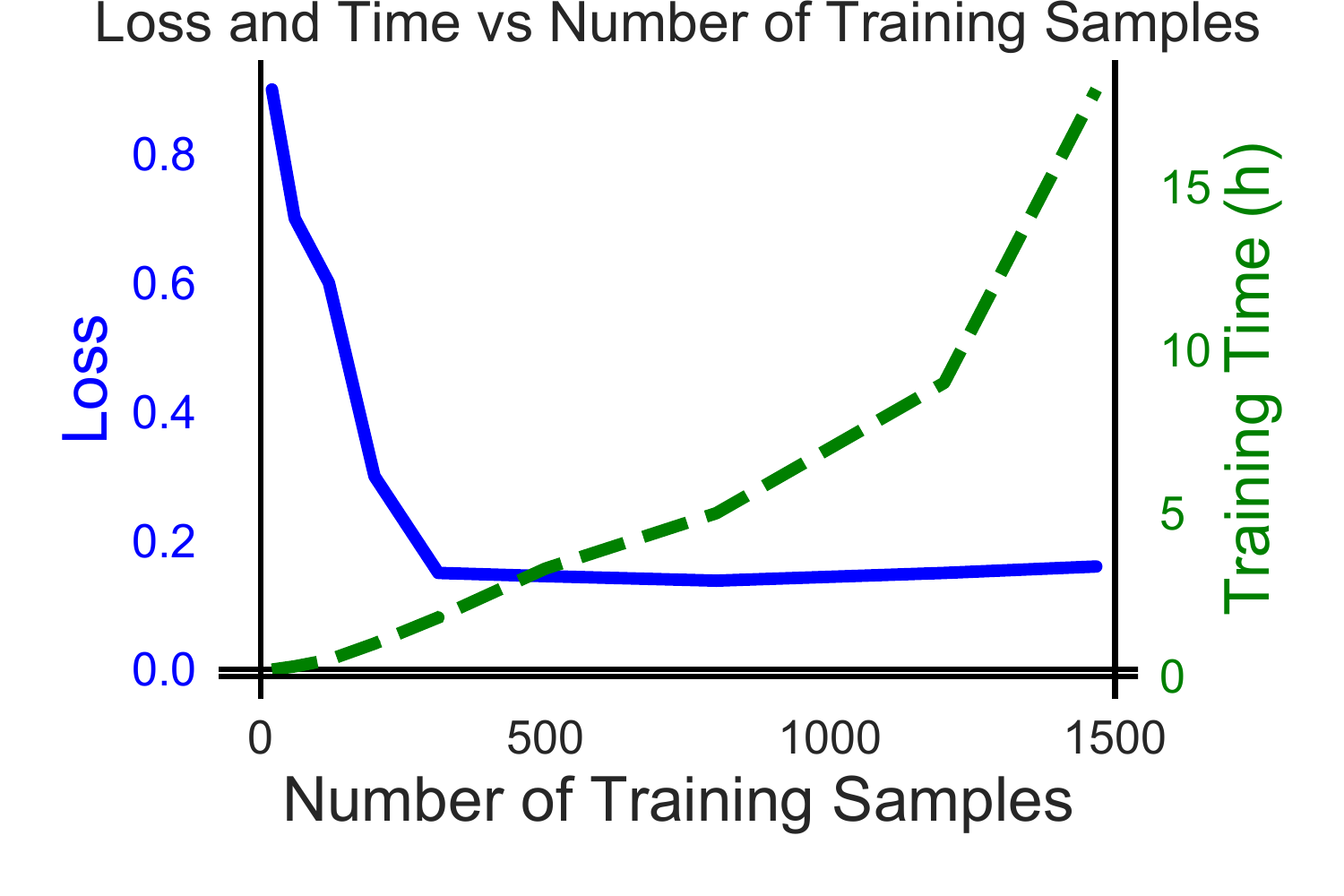}
        \caption{Loss function and training time against the number of training samples used. This was acquired by running a number of experiments using input datasets of different size. Using more than 300 training samples does not give accuracy benefit, while just increasing the training time.}
        \label{fig:results_loss}
    \end{subfigure}
    
    \caption{Evaluation of the transfer learning method using the test dataset in various categories.}
\vspace{-0.5cm}
\end{figure*}

Training was done by using datasets of different sizes containing the Kuka robot. Mini-batches were created in order to make the most out of the available GPU memory and all the data was randomly shuffled to reduce the biases. Before starting any training, parameters for the \textit{frozen} layers were transferred from the old model fully-trained on UR datasets. This ensured that each training had an identical configuration in the beginning. The number of training samples varied by the experiment and the input size of the images was reduced by half from the original dimensions, down to $256\times212$ pixels. The pixel intensity values of the input images were normalised to the range between 0 and 1. The learning rate was set to $0.001$ at the start of the training and then gradually decreased towards $0.000001$ as the training progressed. 

%%% IS THERE SPACE FOR THIS ONE??? %%%

% \begin{figure*}[ht]
% \vspace{0.2cm}
%     \centering
%     \includegraphics[width=0.99\linewidth]{./figures/RobotEstimatedJointPos.pdf}
%     \caption{Estimated robot joint position coordinates marked on the images taken from the dataset. Due to difficulty in visualising 3D coordinates on printed figures, the estimated joint coordinates were mapped back into 2D images. Green crosses indicate the ground truth position, red circles indicate predicted positions of joints and magenta circles indicate the predicted position for the robot base.}
%     \label{fig:marked_coordinates}
% % \vspace{-0.5cm}
% \end{figure*}

% Section Experiments and Results
\section{EXPERIMENTS AND RESULTS}
\label{sec:results}

A number of experiments were carried out in order to determine the effectiveness of the transfer learning process. In order to find the optimum amount of training samples needed for transfer learning, each experiment consisted of a training set with different size, all randomly sampled from the Kuka dataset. The testing set was identical for all the experiments. 

\begin{table}[h]
\caption{Summary of the Transfer Learning results (using 312 samples for training) on the test set of Kuka LBR iiwa robot with a comparison of a Multi-Objective CNN with just Universal Robots.}
\label{table:results_summary}
\centering
%\begin{tabular}{ | c || c | c | c | c | }
\begin{tabular}{ |p{3.1cm}||p{1.5cm}|p{2.2cm}|}
\hline
Measure & Full Training  & Transfer Learning \\
 \hline
 Mask Accuracy, \% & \boldmath$98\%$ & $97.3\%$ \\
 Robot Type Accuracy, \%  & \boldmath$98.3\%$ & --- \\
 Joint Pos Error (Median) & \boldmath$2.46 cm$ & $2.87 cm$ \\
 Base Pos Error (Median) & $2.13 cm$ & \boldmath$2.02 cm$ \\
 Training Time (hours) & $60$ hours & \boldmath$2$ hours \\
 \hline
\end{tabular}
% \vspace{-0.5cm}
\end{table}

The evaluation was done using a testing set by comparing the output against the ground truth data. The robot mask accuracy is defined by counting the number of pixels in the CNN output image that match the ground truth mask. For the robot joint and base coordinates, Euclidean distance between the CNN estimated results and ground truth results was calculated. We compare the results of transfer learning method trained on the Kuka robot against our previously presented multi-objective CNN fully trained for UR robots~\cite{miseikis2018multi}. Results are summarised in Table~\ref{table:results_summary}.

\begin{figure}[ht]
\vspace{0.2cm}
    \centering
    \includegraphics[width=0.98\linewidth]{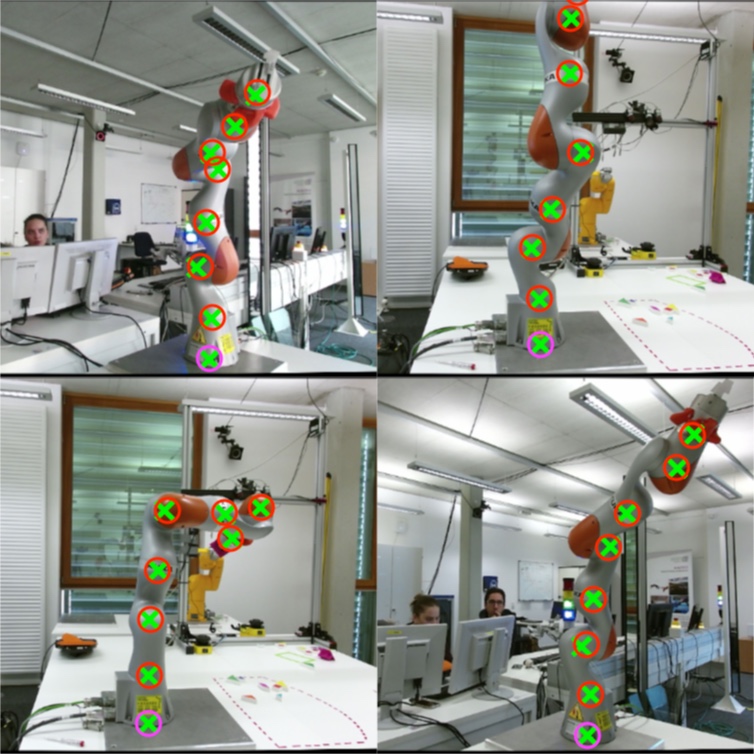}
    \caption{Estimated robot joint position coordinates marked on the images taken from the dataset. Due to difficulty in visualising 3D coordinates on printed figures, the estimated joint coordinates were mapped back into 2D images. Green crosses indicate the ground truth position, red circles indicate predicted positions of joints and magenta circles indicate the predicted position for the robot base. In some cases, even when a part of the robot is out of bounds, positions of visible joints as well as the unseen joint are predicted with centimeter accuracy.}
    \label{fig:marked_coordinates}
\vspace{-0.5cm}
\end{figure}

Compared to a fully trained system, the transfer learning results matched closely. As seen in Figure~\ref{fig:results_all}, the error in estimating 3D positions of robot joints was $2.87$ cm compared to $2.46$ cm in a fully trained system, while the robot mask accuracy difference was just $0.7\%$ with $97.3\%$ for transfer learning method and $98\%$ in a fully trained CNN. Robot base position estimation was actually more accurate in transfer learning method with an error of $2.02$ cm compared to $2.13$ cm. Resulting positions of robot base and joints mapped onto 2D and marked on the example dataset images are shown in Figure~\ref{fig:marked_coordinates}. The system was adapted for just one type of the new robot, so we did not evaluate the accuracy of robot type detection. The system adapted to an additional robot joint in the transfer learning method. There can be seen an increase in the error for Joint 3, and that could be partly caused by a different structure of the robot, as seen in Figure~\ref{fig:results_joints}.

However, the main benefit of the transfer learning method can be seen in training time. By looking at the Figure~\ref{fig:results_loss}, where loss calculation and training time is shown against the number of samples used for training, it can be clearly seen that the optimum result is around 300 training samples. To be specific, it was the experiment, where 312 training samples were used. It took a bit under 2 hours of training and the resulting loss was $0.15$. Having more samples, the loss got down to $0.14$, but it took significantly more time to train. An interesting point was that by using the whole training dataset, the loss increased back to $0.15$ and took around $16$ hours of training. The forward propagation time per sample averaged to $13.5$ ms. All the training and testing was done on Nvidia Geforce GTX 1080 Ti graphics card.

% HOW LONG IS FORWARD-PROPAGATION???

{\color{blue} 
}

\section{CONCLUSIONS AND FUTURE WORK}
\label{sec:conclusion}

In this paper, we have presented a transfer learning approach to adapt a previously trained multi-objective CNN to new types of robots. In general, the system identifies and localises the robot arm and estimates its base and joints' positions in 3D. This allows a camera to be placed in any position and moved around without having to re-calibrate the camera-robot system with Hand-Eye calibration. In this work, we have shown that by taking a fully trained system, a significantly less training data is needed to adapt it for new robot models, which have a different shape, appearance and even more degrees of freedom.

The results have shown that accuracy achieved by using transfer learning closely matches the results of the fully trained system and can even improve in some cases. This means that the system is able to adapt and learn to recognise new robots with having just limited amount of training data. Similarly to what we do when learning new skills and practising them afterwards.

This work can be useful in dynamic environments where it is difficult to predict where robots, sensors and people are located, but operational safety has to be established. By expanding this method to numerous robots, and other equipment, fixed setups and calibration can be discarded. Unfortunately, the accuracy is still in centimetre level, and it is not applicable for precision tasks. However, in many adaptive human-robot and robot-robot interaction tasks, general obstacle avoidance and collaboration movements could be made possible.

Given a precise robot body detection, another possible application could be self-inspection for the robot to detect any unknown and unexpected damage. Similar to new robots are added using transfer learning, typical damages could be taught to the system and identified by the robot scanning itself, observing its own reflection or having another robot to scan it. This can be very useful in environments, like disaster areas, where robots have to work autonomously for long periods of time, or when internal sensors give unusual readings and hull should be inspected.

For future work, we plan to implement more robots as well as having robots on mobile platforms in the system. Instead of training on one new robot model, transfer learning will be used to expand the CNN to work with a line of robots, including the originally trained ones. Previously mentioned robot self-inspection is also of high interest, as well as adding collaborative tasks with people by tracking their movements using the latest skeleton tracking methods. Furthermore, more types of cameras will be tested and transition from one camera to another analysed.

\section*{ACKNOWLEDGMENT}
This work is partially supported by The Research Council of Norway as a part of the Engineering Predictability with Embodied Cognition (EPEC) project, under grant agreement 240862, and by the Austrian Ministry for Transport, Innovation and Technology (BMVIT) within the project framework CollRob (Collaborative Robotics).

\addtolength{\textheight}{-12cm}   % This command serves to balance the column lengths

%%%%%%%%%%%%%%%%%%%%%%%%%%%%%%%%%%%%%%%%%%%%%%%%%%%%%%%%%%%%%%%%%%%%%%%%%%%%%%%%

%%%%%%%%%%%%%%%%%%%%%%%%%%%%%%%%%%%%%%%%%%%%%%%%%%%%%%%%%%%%%%%%%%%%%%%%%%%%%%%%

%%%%%%%%%%%%%%%%%%%%%%%%%%%%%%%%%%%%%%%%%%%%%%%%%%%%%%%%%%%%%%%%%%%%%%%%%%%%%%%%
%\section*{APPENDIX}

%\section*{ACKNOWLEDGMENT}

%%%%%%%%%%%%%%%%%%%%%%%%%%%%%%%%%%%%%%%%%%%%%%%%%%%%%%%%%%%%%%%%%%%%%%%%%%%%%%%%

\bibliographystyle{IEEEtran}
\bibliography{IEEEexample}

\begin{thebibliography}{10}
\providecommand{\url}[1]{#1}
\csname url@rmstyle\endcsname
\providecommand{\newblock}{\relax}
\providecommand{\bibinfo}[2]{#2}
\providecommand\BIBentrySTDinterwordspacing{\spaceskip=0pt\relax}
\providecommand\BIBentryALTinterwordstretchfactor{4}
\providecommand\BIBentryALTinterwordspacing{\spaceskip=\fontdimen2\font plus
\BIBentryALTinterwordstretchfactor\fontdimen3\font minus
  \fontdimen4\font\relax}
\providecommand\BIBforeignlanguage[2]{{%
\expandafter\ifx\csname l@#1\endcsname\relax
\typeout{** WARNING: IEEEtran.bst: No hyphenation pattern has been}%
\typeout{** loaded for the language `#1'. Using the pattern for}%
\typeout{** the default language instead.}%
\else
\language=\csname l@#1\endcsname
\fi
#2}}

\bibitem{lee2015cyber}
J.~Lee, B.~Bagheri, and H.-A. Kao, ``A cyber-physical systems architecture for
  industry 4.0-based manufacturing systems,'' \emph{Manufacturing Letters},
  vol.~3, pp. 18--23, 2015.

\bibitem{krizhevsky2012imagenet}
A.~Krizhevsky, I.~Sutskever, and G.~E. Hinton, ``Imagenet classification with
  deep convolutional neural networks,'' in \emph{Advances in neural information
  processing systems}, 2012, pp. 1097--1105.

\bibitem{oquab2014learning}
M.~Oquab, L.~Bottou, I.~Laptev, and J.~Sivic, ``Learning and transferring
  mid-level image representations using convolutional neural networks,'' in
  \emph{Proceedings of the IEEE conference on computer vision and pattern
  recognition}, 2014, pp. 1717--1724.

\bibitem{greenspan2016guest}
H.~Greenspan, B.~van Ginneken, and R.~M. Summers, ``Guest editorial deep
  learning in medical imaging: Overview and future promise of an exciting new
  technique,'' \emph{IEEE Transactions on Medical Imaging}, vol.~35, no.~5, pp.
  1153--1159, 2016.

\bibitem{ng2015deep}
H.-W. Ng, V.~D. Nguyen, V.~Vonikakis, and S.~Winkler, ``Deep learning for
  emotion recognition on small datasets using transfer learning,'' in
  \emph{Proceedings of the 2015 ACM on international conference on multimodal
  interaction}.\hskip 1em plus 0.5em minus 0.4em\relax ACM, 2015, pp. 443--449.

\bibitem{xie2015transfer}
M.~Xie, N.~Jean, M.~Burke, D.~Lobell, and S.~Ermon, ``Transfer learning from
  deep features for remote sensing and poverty mapping,'' \emph{arXiv preprint
  arXiv:1510.00098}, 2015.

\bibitem{shin2016deep}
H.-C. Shin, H.~R. Roth, M.~Gao, L.~Lu, Z.~Xu, I.~Nogues, J.~Yao, D.~Mollura,
  and R.~M. Summers, ``Deep convolutional neural networks for computer-aided
  detection: Cnn architectures, dataset characteristics and transfer
  learning,'' \emph{IEEE transactions on medical imaging}, vol.~35, no.~5, pp.
  1285--1298, 2016.

\bibitem{weiss2016survey}
K.~Weiss, T.~M. Khoshgoftaar, and D.~Wang, ``A survey of transfer learning,''
  \emph{Journal of Big Data}, vol.~3, no.~1, p.~9, 2016.

\bibitem{miseikis2018multi}
J.~Miseikis, I.~Brijacak, S.~Yahyanejad, K.~Glette, O.~J. Elle, and
  J.~Torresen, ``Multi-objective convolutional neural networks for robot
  localisation and 3d position estimation in 2d camera images,'' \emph{arXiv
  preprint arXiv:1804.03005}, 2018.

\bibitem{mainprice2013human}
J.~Mainprice and D.~Berenson, ``Human-robot collaborative manipulation planning
  using early prediction of human motion,'' in \emph{Intelligent Robots and
  Systems (IROS), 2013 IEEE/RSJ International Conference on}.\hskip 1em plus
  0.5em minus 0.4em\relax IEEE, 2013, pp. 299--306.

\bibitem{miseikis2016multi}
J.~Mi{\v{s}}eikis, K.~Glette, O.~J. Elle, and J.~Torresen, ``{Multi 3D camera
  mapping for predictive and reflexive robot manipulator trajectory
  estimation},'' in \emph{Computational Intelligence (SSCI), 2016 IEEE
  Symposium Series on}.\hskip 1em plus 0.5em minus 0.4em\relax IEEE, 2016, pp.
  1--8.

\bibitem{miseikis2016automatic}
J.~Miseikis, K.~Glette, O.~J. Elle, and J.~Torresen, ``Automatic calibration of
  a robot manipulator and multi 3d camera system,'' in \emph{System Integration
  (SII), 2016 IEEE/SICE International Symposium on}.\hskip 1em plus 0.5em minus
  0.4em\relax IEEE, 2016, pp. 735--741.

\bibitem{Fankhauser2015KinectV2ForMobileRobotNavigation}
P.~Fankhauser, M.~Bloesch, D.~Rodriguez, , R.~Kaestner, M.~Hutter, and
  R.~Siegwart, ``Kinect v2 for mobile robot navigation: Evaluation and
  modeling,'' in \emph{IEEE International Conference on Advanced Robotics
  (ICAR) (submitted)}, 2015.

\bibitem{heikkila2000flexible}
T.~Heikkil{\"a}, M.~Sallinen, T.~Matsushita, and F.~Tomita, ``Flexible hand-eye
  calibration for multi-camera systems,'' in \emph{Intelligent Robots and
  Systems, 2000.(IROS 2000). Proceedings. 2000 IEEE/RSJ International
  Conference on}, vol.~3.\hskip 1em plus 0.5em minus 0.4em\relax IEEE, 2000,
  pp. 2292--2297.

\bibitem{sucan2013moveit}
I.~A. Sucan and S.~Chitta, ``{MoveIt!}'' \emph{Online Available:
  http://moveit.ros.org}, 2013.

\bibitem{meeussen2012urdf}
W.~Meeussen, J.~Hsu, and R.~Diankov, ``Urdf-unified robot description format,''
  2012.

\end{thebibliography}

\end{document}